\documentclass[journal]{IEEEtran}

\usepackage{float}
\usepackage{array}

\usepackage{amsmath}
\usepackage{amssymb}
\ifCLASSOPTIONcompsoc
  \usepackage[caption=false,font=normalsize,labelfont=sf,textfont=sf]{subfig}
\else
  \usepackage[caption=false,font=footnotesize]{subfig}
\fi

\usepackage{multirow}
\usepackage{url}
\usepackage{microtype}
\usepackage{graphicx}
\usepackage{xcolor}
\usepackage{tikz}
\usetikzlibrary{arrows.meta,positioning,fit,backgrounds}
\graphicspath{{./}{/}}

\definecolor{TextDark}{HTML}{2E2E2E}
\definecolor{TextLight}{HTML}{666666}
\definecolor{MorandiGray}{HTML}{B8B8B8}
\definecolor{MorandiBlue}{HTML}{7BAFD4}
\definecolor{MorandiTeal}{HTML}{6FB3A8}
\definecolor{MorandiPink}{HTML}{D9A1A9}
\definecolor{MorandiGreen}{HTML}{9ACD9A}
\definecolor{MorandiPurple}{HTML}{B2A1C7}

\usetikzlibrary{arrows.meta,positioning,fit,backgrounds,calc,shapes.geometric}

\tikzset{
  decision/.style={
    diamond, aspect=2.5,        
    draw=MorandiBlue!80!black,
    fill=MorandiBlue!10,
    minimum width=2.6cm,
    minimum height=1.2cm,
    align=center,
    inner sep=2pt,
    text=TextDark,
    font=\small
  }
}

\definecolor{MorandiBlue}{RGB}{150,165,180}
\definecolor{MorandiGreen}{RGB}{165,175,155}
\definecolor{MorandiPink}{RGB}{200,185,175}
\definecolor{MorandiBeige}{RGB}{220,210,195}
\definecolor{MorandiGray}{RGB}{180,175,170}
\definecolor{MorandiBrown}{RGB}{185,170,155}
\definecolor{MorandiPurple}{RGB}{175,165,180}
\definecolor{MorandiTeal}{RGB}{160,175,170}
\definecolor{TextDark}{RGB}{60,60,60}
\definecolor{TextLight}{RGB}{120,120,120}

\begin{document}

\title{ST-DETrack: Identity-Preserving Branch Tracking in Entangled Plant Canopies via Dual Spatiotemporal Evidence}

\author{Yueqianji~Chen,
        Kevin~Williams,
        John~H.~Doonan,
        Paolo~Remagnino$^{*}$,~\IEEEmembership{Senior~Member,~IEEE,}
        and~Jo~Hepworth$^{*}$
\thanks{Y. Chen (Formerly) J. Hepworth, Department of Biosciences, Durham University, Stockton Road, Durham DH1 3LE, U.K.}%
\thanks{P. Remagnino, Department of Computer Science, Durham University, Stockton Road, Durham DH1 3LE, U.K.}%
\thanks{K. Williams, J. H. Doonan, National Plant Phenomics Centre, IBERS, Aberystwyth University, Aberystwyth SY23 3EE, U.K.}%
\thanks{$^{*}$Corresponding authors: Paolo Remagnino and Jo Hepworth.}}

\markboth{IEEE Transactions on Image Processing,~Vol.~XX, No.~X, Month~2025}%
{Chen \MakeLowercase{\textit{et al.}}: ST-DETrack: Separating Entangled Plant Branches via Dual-Evidence Spatio-Temporal Association}

\maketitle

\begin{abstract}
Automated extraction of individual plant branches from time-series imagery is essential for high-throughput phenotyping, yet it remains computationally challenging due to non-rigid growth dynamics and severe identity fragmentation within entangled canopies. To overcome these stage-dependent ambiguities, we propose ST-DETrack, a spatiotemporal-fusion dual-decoder network designed to preserve branch identity from budding to flowering. Our architecture integrates a spatial decoder, which leverages geometric priors such as position and angle for early-stage tracking, with a temporal decoder that exploits motion consistency to resolve late-stage occlusions. Crucially, an adaptive gating mechanism dynamically shifts reliance between these spatial and temporal cues, while a biological constraint based on negative gravitropism mitigates vertical growth ambiguities. Validated on a \textit{Brassica napus} dataset, ST-DETrack achieves a Branch Matching Accuracy (BMA) of 93.6\%, significantly outperforming spatial and temporal baselines by 28.9 and 3.3 percentage points, respectively. These results demonstrate the method's robustness in maintaining long-term identity consistency amidst complex, dynamic plant architectures.
\end{abstract}


\IEEEpeerreviewmaketitle

\section{Introduction}
\label{sec:intro}

\IEEEPARstart{T}{he} automated extraction of individual topological structures from time-series imagery is a fundamental challenge in computer vision, with critical applications ranging from medical vessel tracking to agricultural phenotyping. In the domain of plant functional genomics, the precise quantification of branching architecture—defined by branch number, angle, and spatial distribution—is essential for understanding yield formation and canopy light interception~\cite{lei2024branch}. While genomic studies require analyzing thousands of plants, manual measurement is destructive and labor-intensive, necessitating robust automated solutions.

The core computational problem lies in the \textit{spatiotemporal segmentation of evolving deformable structures}. Unlike pedestrian tracking or vehicle re-identification tasks where objects maintain rigid shapes and coherent motion, plant branches exhibit complex, non-rigid growth dynamics. Specifically, the task involves associating terminal organs (e.g., floral buds) with their originating sites (branch points) to reconstruct independent branch instances. This problem presents a unique ``\textit{stage-dependent ambiguity}'' that defies conventional single-modality approaches.

\begin{figure}[ht]
  \centering
  \includegraphics[width=1\columnwidth]{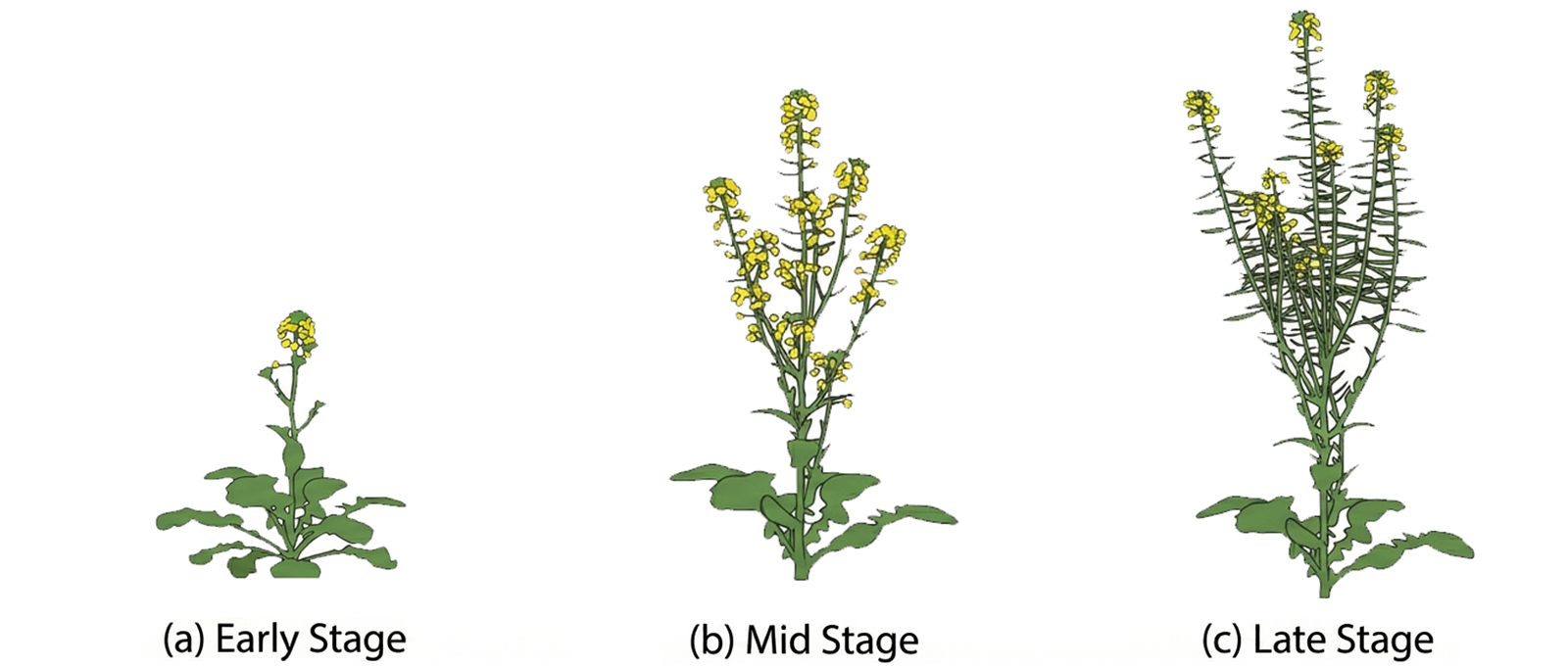}
  \caption{Morphological evolution of rapeseed branching architecture across growth stages. Spatial ambiguity increases sharply as branches elongate and intertwine, necessitating temporal evidence for reliable identity assignment.}
  \label{fig:growth_stages}
\end{figure}  

As illustrated in Fig.~\ref{fig:growth_stages}, the morphological evolution of plants creates a shifting evidential landscape. At early developmental stages, branches are sparse and spatially distinct. In this regime, geometric priors—such as Euclidean proximity and angular alignment—are sufficient for determining topological connectivity. However, as plants mature into late flowering and seed-set stages, branches elongate and intertwine, creating a dense ``spaghetti-like'' canopy. Severe occlusions and spatial overlaps emerge, where multiple buds from different branches may occupy adjacent spatial coordinates. Under such conditions, spatial geometry becomes ambiguous, and the identity of a branch can only be resolved by exploiting temporal continuity from historical frames.

Existing literature has largely tackled parts of this problem in isolation, broadly categorized into static instance segmentation and multi-object tracking (MOT).

\textit{Static Instance Segmentation Limitations:} Deep learning-based segmentation frameworks have become the de facto standard for extracting plant organs. Two-stage detectors like Mask R-CNN~\cite{he2017mask} have been extensively adapted for segmenting leaves, fruits, and stems by generating region proposals followed by pixel-level classification~\cite{ubbens2018use,casado2020stomatal}. Conversely, single-stage models such as YOLOv8-Seg~\cite{ultralytics_yolov8} offer real-time efficiency and have been applied to high-throughput crop detection~\cite{tao2024branch}. While these methods achieve high per-frame accuracy in sparse scenes, they treat each frame independently. In longitudinal sequences, this independence leads to \textit{identity fragmentation}: a single growing branch is often assigned inconsistent IDs across frames due to minor appearance changes or temporary occlusions, rendering the reconstruction of developmental trajectories impossible. Our previous work, PBG-Net, attempted to impose spatial constraints via junction detection, but similarly failed to resolve identities in dense, entangled scenarios where spatial cues are inherently ambiguous.

\textit{Multi-Object Tracking Limitations:} To enforce temporal consistency, tracking-by-detection paradigms are often employed. Classical approaches like DeepSORT~\cite{wojke2017simple} combine Kalman filtering with appearance embeddings. However, these methods rely on inertial motion assumptions (e.g., constant velocity) that hold for pedestrians but are violated by biological growth. Plants exhibit non-rigid expansion driven by biological priors (e.g., gravitropism and phototropism) rather than momentum. Furthermore, transformer-based trackers like TrackFormer~\cite{meinhardt2022trackformer} or TransTrack~\cite{sun2020transtrack}, which propagate queries across frames, typically assume a fixed object topology over short windows. They lack the specialized mechanisms to handle the week-long appearance drift and the emergence of new organs characteristic of plant development.

To address these limitations, we hypothesize that \textit{the optimal strategy for segmenting growing structures is to dynamically re-weight spatial and temporal evidence based on structural maturity.} We propose a \textbf{Spatiotemporal-Fusion Dual-Decoder Network} that disentangles these two evidence streams.
The architecture features two specialized decoders: a \textit{Spatial Decoder} that leverages geometric compatibility (position, angle) to resolve topology in early stages, and a \textit{Temporal Decoder} that exploits motion consistency and a novel \textit{negative gravitropism prior} to maintain identities through occlusion in late stages. An adaptive gating mechanism automatically learns to shift reliance from spatial to temporal cues as the canopy complexity increases. Finally, to convert discrete point matches into continuous morphological descriptions, we employ B-spline fitting on the skeletal graph to generate pixel-perfect branch segmentation.

In summary, our main contributions are:
\begin{itemize}
    \item We formalize the extraction of individual plant branches as a spatiotemporal matching problem characterized by stage-dependent ambiguity, generalizing beyond specific crop types.
    \item We propose a dual-decoder architecture with an adaptive fusion mechanism that explicitly models the trade-off between geometric spatial priors and biological motion priors.
    \item We introduce a biological constraint based on negative gravitropism to refine temporal association, significantly reducing errors in vertical growth tracking.
    \item Extensive experiments on a large-scale \textit{Brassica napus} dataset demonstrate that our method outperforms state-of-the-art segmentation (Mask R-CNN) and tracking (DeepSORT) baselines, particularly in handling complex, entangled canopies.
\end{itemize}

\section{Methodology}
\label{sec:method}

\subsection{Problem Formulation}
\label{sec:problem_formulation}

We formalize the task of individual branch extraction as a \textit{spatiotemporal matching problem} over time-series side-view images $\{I_t\}_{t=1}^T$. At each time point $t$, the image $I_t \in \mathbb{R}^{H \times W \times 3}$ contains two key anatomical structures:
\begin{itemize}
    \item \textbf{Branch points} $\mathcal{B}_t = \{b_j^t\}_{j=1}^{K_t}$: Junctions where primary lateral branches attach to the main stem, defined as $b_j^t = (\text{order}_j, x_j, y_j, \theta_j) \in \mathbb{R}^4$, encoding spatial position $(x_j, y_j)$, branch orientation $\theta_j$, and botanical rank $\text{order}_j$.
    \item \textbf{Floral buds} $\mathcal{F}_t = \{f_i^t\}_{i=1}^{M_t}$: Terminal endpoints of branches, represented as $f_i^t = (\text{order}_i, x_i, y_i, w_i, h_i) \in \mathbb{R}^5$, with bounding box $(x_i, y_i, w_i, h_i)$ and inherited branch order $\text{order}_i$.
\end{itemize}

\noindent\textbf{Goal.} Establish a one-to-one correspondence $\pi_t: \mathcal{F}_t \rightarrow \mathcal{B}_t$ that assigns each bud to its originating branch point, enabling extraction of independent primary branch instances. This mapping must satisfy:
\begin{equation}
\pi_t(f_i) = b_j \quad \Leftrightarrow \quad \text{bud } f_i \text{ attaches to branch } b_j
\end{equation}

\noindent\textbf{Challenges.} Two key ambiguities complicate $\pi_t$ inference:
\begin{enumerate}
    \item \textbf{Spatial ambiguity}: In dense canopies (late flowering stages), multiple buds from different branches may reside at similar distances from a given branch point $b_j$, rendering proximity-based matching unreliable. Formally, $\exists f_i, f_k \in \mathcal{F}_t, i \neq k$ such that $\|\mathbf{p}(f_i) - \mathbf{p}(b_j)\| \approx \|\mathbf{p}(f_k) - \mathbf{p}(b_j)\|$, where $\mathbf{p}(\cdot)$ denotes position.
    \item \textbf{Temporal ambiguity}: Transient occlusions caused by overlapping branches may cause buds to disappear in frame $t$ and reappear in $t+1$, disrupting trajectory continuity. Moreover, new buds emerge continuously as branches elongate, lacking historical correspondence.
\end{enumerate}

These challenges necessitate a hybrid approach that dynamically weights spatial evidence (geometric consistency within frame $t$) and temporal evidence (motion continuity from $t-1$ to $t$) based on scene complexity and branch maturity.

\subsection{Overall Framework}
\label{sec:overall_framework}

\begin{figure*}[t]
    \centering
    \includegraphics[width=1\textwidth]{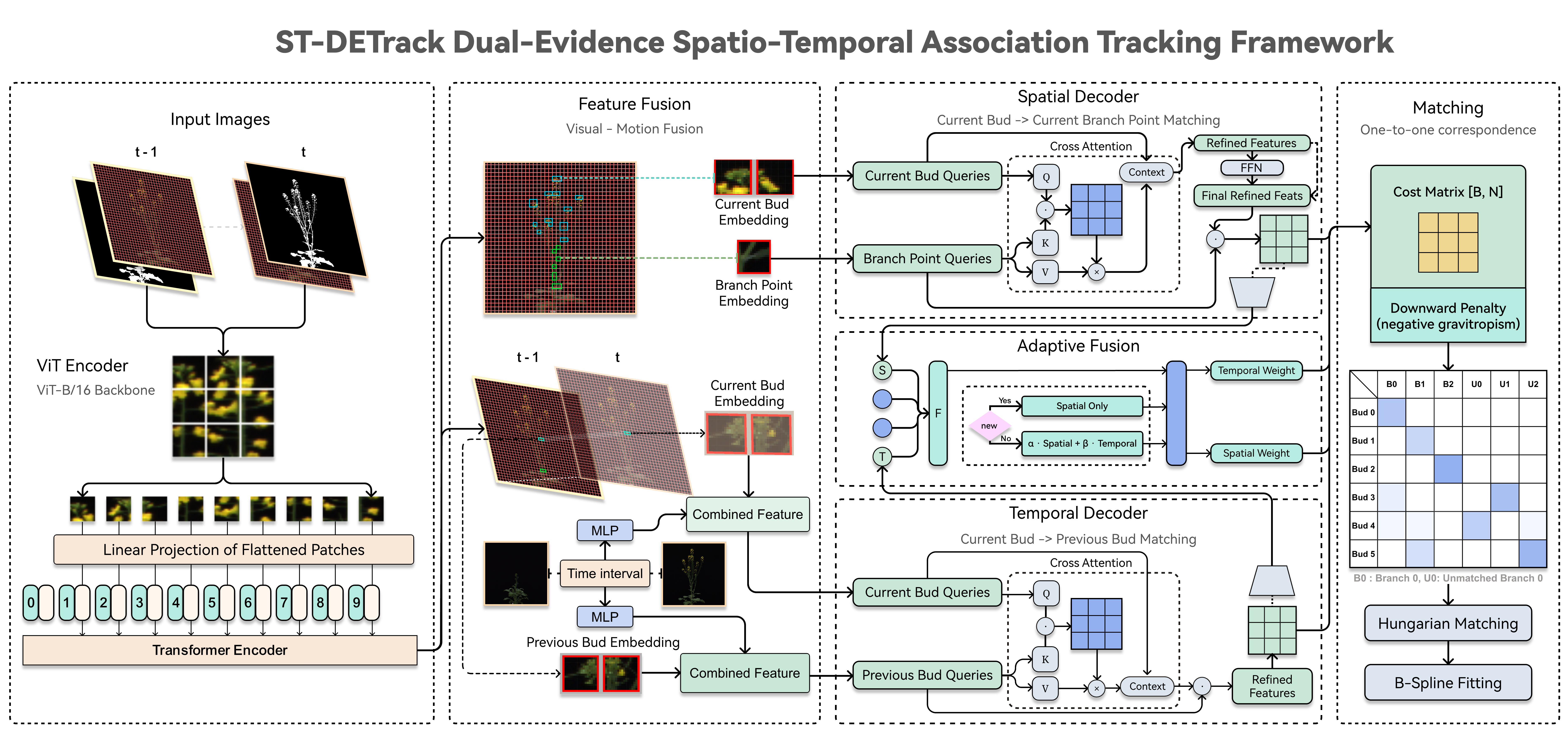}
    \caption{Overall dual-decoder fusion architecture comprising a shared encoder, spatial decoder, temporal decoder, and adaptive fusion mechanism.}
    \label{fig:fusion_architecture}
\end{figure*}

Figure~\ref{fig:fusion_architecture} illustrates the proposed spatiotemporal-fusion dual-decoder architecture, which explicitly models and combines both evidence streams through learnable adaptive gating.

\noindent\textbf{Input Representation.} For each frame pair $(I_{t-1}, I_t)$, we construct:
\begin{itemize}
    \item \textbf{Concatenated image} $\tilde{I}_t \in \mathbb{R}^{H \times W \times 8}$: Stack previous RGB + mask ($I_{t-1}$) and current RGB + mask ($I_t$), where masks indicate plant foreground to suppress background noise.
    \item \textbf{Motion-augmented buds} $\mathcal{F}_t^{\text{aug}} = \{(f_i^t, \mathbf{m}_i^{t-1})\}$: Each current bud $f_i^t$ is augmented with historical motion features $\mathbf{m}_i^{t-1} = (x_i^{t-1}, y_i^{t-1}, v_{x,i}, v_{y,i}, a_{x,i}, a_{y,i}) \in \mathbb{R}^6$, encoding previous position, velocity, and acceleration computed via finite differences or Kalman filtering.
\end{itemize}

\noindent\textbf{Shared Encoder.} A Vision Transformer (ViT-B/16) with modified patch embedding layer processes 8-channel input:
\begin{equation}
f_t = \text{ViT}_{\text{enc}}(\tilde{I}_t) \in \mathbb{R}^{B \times 768 \times 14 \times 14}
\end{equation}
The encoder produces spatially structured feature maps preserving geometric relationships essential for position-sensitive matching.

\noindent\textbf{Dual Decoders.} Two parallel decoder heads process $f_t$ with complementary conditioning:
\begin{enumerate}
    \item \textbf{Spatial Decoder} $\mathcal{D}_{\text{spatial}}$: Outputs current-frame matching scores $M^{\text{spatial}} \in \mathbb{R}^{|\mathcal{F}_t| \times |\mathcal{B}_t|}$ based on geometric compatibility (Section~\ref{sec:spatial_module}).
    \item \textbf{Temporal Decoder} $\mathcal{D}_{\text{temporal}}$: Outputs cross-frame association scores $M^{\text{temporal}} \in \mathbb{R}^{|\mathcal{F}_t| \times |\mathcal{F}_{t-1}|}$ leveraging motion priors (Section~\ref{sec:temporal_module}).
\end{enumerate}

\noindent\textbf{Fusion Output.} An adaptive gating mechanism (Section~\ref{sec:fusion_mechanism}) produces per-branch weights $w^{\text{spatial}}_j, w^{\text{temporal}}_j$ to combine evidence streams into the final matching matrix $M^{\text{fusion}} \in \mathbb{R}^{|\mathcal{F}_t| \times |\mathcal{B}_t|}$. Hungarian algorithm enforces one-to-one assignment at inference.

\subsection{Spatial Evidence Module}
\label{sec:spatial_module}

The spatial decoder captures geometric relationships within the current frame, mimicking expert visual reasoning: "Is this bud spatially consistent with originating from branch point $b_j$?"

\subsubsection{Geometric Feature Encoding}

For each bud-branch pair $(f_i, b_j)$, we extract four complementary geometric descriptors:
\begin{enumerate}
    \item \textbf{Relative position}: $\Delta \mathbf{p}_{ij} = (x_i - x_j, y_i - y_j) \in \mathbb{R}^2$, capturing horizontal/vertical displacement.
    \item \textbf{Euclidean distance}: $d_{ij} = \|\Delta \mathbf{p}_{ij}\|_2$, normalized by image diagonal to ensure scale invariance.
    \item \textbf{Angular alignment}: $\alpha_{ij} = \angle(\Delta \mathbf{p}_{ij}, \mathbf{d}_j)$, measuring deviation between bud direction and branch orientation $\theta_j$ (encoded as unit vector $\mathbf{d}_j = (\cos\theta_j, \sin\theta_j)$).
    \item \textbf{Bounding box geometry}: Aspect ratio $w_i/h_i$ and area $w_i \cdot h_i$ as proxies for bud maturity.
\end{enumerate}
These cues are summarized into a low-dimensional geometric feature vector $\mathbf{g}_{ij}$ (containing normalized positions, bounding-box geometry, and branch orientation) and embedded via MLP: $\mathbf{e}_{\text{geom}} = \text{MLP}_{\text{geom}}(\mathbf{g}_{ij}) \in \mathbb{R}^{512}$.

\subsubsection{Visual Feature Integration}

Beyond geometry, the decoder samples visual features from encoder output $f_t$:
\begin{itemize}
    \item \textbf{Bud visual features}: Sample $f_t$ at bud center $(x_i, y_i)$ and project to hidden dimension: $\mathbf{v}_{\text{bud}} = W_{\text{vis}} \cdot f_t[:, y_i', x_i'] \in \mathbb{R}^{512}$, where $(x_i', y_i')$ are grid-normalized coordinates.
    \item \textbf{Branch visual features}: To capture directional context, sample along branch orientation $\theta_j$ at multiple offsets $\delta \in \{0, 0.05, 0.10\}$ (fraction of image diagonal), then average:
\begin{equation}
\begin{split}
\mathbf{v}_{\text{branch}} = \frac{1}{|\Delta|} \sum_{\delta \in \Delta} W_{\text{vis}} \cdot f_t[:, (y_j + \delta \sin\theta_j)', \\
\qquad (x_j + \delta \cos\theta_j)']
\end{split}
\end{equation}
    This multi-scale sampling approximates branch appearance along its trajectory.
\end{itemize}

\subsubsection{Cross-Attention Matching}

Bud and branch features are integrated via multi-head cross-attention~\cite{vaswani2017attention}:
\begin{align}
\text{Query:} \quad Q_{\text{bud}} &= [\mathbf{e}_{\text{geom}}^{(1)}, \ldots, \mathbf{e}_{\text{geom}}^{(M)}] \notag \\
&\quad + [\mathbf{v}_{\text{bud}}^{(1)}, \ldots, \mathbf{v}_{\text{bud}}^{(M)}] \in \mathbb{R}^{M \times 512} \\
\text{Key/Value:} \quad K_{\text{branch}}, V_{\text{branch}} &= [\mathbf{v}_{\text{branch}}^{(1)}, \ldots, \mathbf{v}_{\text{branch}}^{(K)}] \in \mathbb{R}^{K \times 512}
\end{align}
Attention output $\text{Attn}(Q, K, V)$ aggregates branch-context information for each bud. A final feed-forward network refines representations, followed by dot-product similarity to compute logits:
\begin{equation}
M^{\text{spatial}}_{ij} = \frac{1}{\tau_{\text{spatial}}} \langle \text{FFN}(Q_i), V_j \rangle
\end{equation}
where $\tau_{\text{spatial}} = 1.0$ is a temperature hyperparameter controlling logit sharpness.

\subsection{Temporal Evidence Module}
\label{sec:temporal_module}

The temporal decoder addresses spatial ambiguities by tracking bud identities across frames, leveraging motion priors and appearance consistency.

\subsubsection{Motion Feature Representation}

For each historical bud $f_i^{t-1}$, we compute a kinematic state:
\begin{equation}
\mathbf{m}_i^{t-1} = (x_i^{t-1}, y_i^{t-1}, v_{x,i}, v_{y,i}, a_{x,i}, a_{y,i})
\end{equation}
Velocity $\mathbf{v}_i = (v_{x,i}, v_{y,i})$ and acceleration $\mathbf{a}_i = (a_{x,i}, a_{y,i})$ are computed via finite differences over a 3-frame window when available, or set to zero for newly appeared buds. This motion representation enables trajectory prediction:
\begin{equation}
\hat{\mathbf{p}}_i^t = \mathbf{p}_i^{t-1} + \mathbf{v}_i \Delta t + \frac{1}{2} \mathbf{a}_i (\Delta t)^2
\end{equation}
where $\Delta t$ is the time interval (in days) between frames and is encoded via a time-interval embedding. This constant-acceleration formulation defines a trajectory prior that the network can exploit when combining motion features with the learned embeddings.

\subsubsection{Temporal Feature Embedding}

The decoder constructs separate embeddings for current and previous buds:
\begin{align}
\mathbf{e}_{\text{curr}}^{(i)} &= \text{MLP}_{\text{curr}}(\mathbf{p}_i^t) + W_{\text{vis}} f_t[:, y_i', x_i'] \notag \\
&\quad + \mathbf{r}_{\text{curr}} + \mathbf{t}_{\Delta t} \\
\mathbf{e}_{\text{prev}}^{(k)} &= \text{MLP}_{\text{prev}}(\mathbf{p}_k^{t-1}, \mathbf{m}_k^{t-2}) + W_{\text{vis}} f_t[:, y_k', x_k'] \notag \\
&\quad + \mathbf{r}_{\text{prev}} + \mathbf{t}_{\Delta t}
\end{align}
where:
\begin{itemize}
    \item $\text{MLP}_{\text{curr}}, \text{MLP}_{\text{prev}}$: Separate 2-layer MLPs processing 10-dimensional inputs comprising position, bounding-box geometry, and (for historical buds) motion features.
    \item $\mathbf{r}_{\text{curr}}, \mathbf{r}_{\text{prev}} \in \mathbb{R}^{512}$: Learnable role embeddings distinguishing temporal roles (``query bud at $t$'' vs.\ ``reference bud at $t-1$'').
    \item $\mathbf{t}_{\Delta t} = \text{MLP}_{\text{time}}([\Delta t])$: Time-interval encoding accounting for variable frame spacing (e.g., daily vs.\ weekly imaging).
\end{itemize}

\subsubsection{Cross-Frame Attention}

Current buds attend to previous buds via multi-head cross-attention, enabling association based on appearance and predicted motion:
\begin{align}
Q_{\text{curr}} &= \text{LayerNorm}([\mathbf{e}_{\text{curr}}^{(1)}, \ldots, \mathbf{e}_{\text{curr}}^{(M)}]) \in \mathbb{R}^{M \times 512} \\
K_{\text{prev}}, V_{\text{prev}} &= \text{LayerNorm}([\mathbf{e}_{\text{prev}}^{(1)}, \ldots, \mathbf{e}_{\text{prev}}^{(N)}]) \in \mathbb{R}^{N \times 512} \\
Z_{\text{cross}} &= \text{CrossAttn}(Q_{\text{curr}}, K_{\text{prev}}, V_{\text{prev}})
\end{align}
Refined features $Z_{\text{cross}}$ are further processed via FFN and residual connections. Matching scores are computed via scaled dot-product:
\begin{equation}
M^{\text{temporal}}_{ik} = \frac{1}{\tau_{\text{temporal}}} \langle \text{FFN}(Z_{\text{cross}}^{(i)}), V_{\text{prev}}^{(k)} \rangle
\end{equation}
where $\tau_{\text{temporal}} = 1.2$ provides slightly softer scores than spatial matching ($\tau = 1.0$). This elevated temperature increases the entropy of the matching distribution, accounting for feature drift caused by plant growth and deformation between frames, thereby preventing overconfident assignments when temporal correspondence is ambiguous.

\subsubsection{Trajectory-to-Branch Conversion}

The temporal decoder outputs bud-to-bud correspondences $M^{\text{temporal}} \in \mathbb{R}^{M_t \times N_{t-1}}$. To enable fusion with spatial evidence, we convert this to bud-to-branch scores $M^{\text{temporal} \to \text{branch}} \in \mathbb{R}^{M_t \times K_t}$ via \textit{order-based aggregation}:
\begin{equation}
M^{\text{temporal} \to \text{branch}}_{ij} =
\begin{cases}
\text{LSE}_k M^{\text{temporal}}_{ik} - \log |\mathcal{S}_j| & \mathcal{S}_j \neq \emptyset \\
\beta_{\text{absent}} & \text{otherwise}
\end{cases}
\label{eq:temporal_to_branch}
\end{equation}
where $\mathcal{S}_j$ denotes previous buds sharing branch $b_j$'s order. Formally, $\mathcal{S}_j = \{k \in \mathcal{F}_{t-1} : \text{order}_k = \text{order}_j\}$, $\text{LSE}$ denotes LogSumExp, and the bias $\beta_{\text{absent}} = -6.0$ penalizes branches without historical evidence. This value corresponds to a prior probability of $e^{-6} \approx 0.25\%$, establishing a strict threshold within the LogSumExp aggregation that effectively filters false positives by requiring accumulated temporal evidence to significantly exceed this noise floor.

\textbf{Geometric Refinement via Negative Gravitropism Prior:} Plant branches exhibit negative gravitropism (upward growth against gravity), providing a strong biological prior for temporal matching. We estimate the global vertical displacement $\Delta y_{\text{global}}$ as the $y$-coordinate shift of the topmost buds between frames. This captures the plant's overall upward growth trend. Matches that violate this negative gravitropic tendency are penalized:
\begin{equation}
\begin{aligned}
M^{\text{temporal} \to \text{branch}}_{ij} &\leftarrow \\
& M^{\text{temporal} \to \text{branch}}_{ij} - \lambda_{\text{vert}} \\
&\times \max(0, |\Delta y_{ik} - \Delta y_{\text{global}}| - \epsilon_{\text{tol}})
\end{aligned}
\end{equation}
where $\Delta y_{ik} = y_i^t - y_k^{t-1}$ for $k \in \mathcal{S}_j$, $\lambda_{\text{vert}} = 6.0$ weights the gravitropic constraint, and $\epsilon_{\text{tol}} = 0.0$ (tolerance threshold). This constraint encodes the biological expectation that all buds on a plant should exhibit coordinated upward displacement driven by apical dominance and auxin-mediated growth regulation.

\subsubsection{Strengths and Limitations}

\textbf{Strengths:} Robust to severe spatial overlap and occlusion by exploiting temporal continuity. Motion prediction via velocity/acceleration provides strong priors for association even when appearance changes.


\subsection{Adaptive Fusion Mechanism}
\label{sec:fusion_mechanism}

The fusion module dynamically combines spatial and temporal evidence based on branch-specific context, addressing the complementary failure modes of each decoder.

\subsubsection{Branch-Dependent Gating}

For each branch $b_j$, we compute fusion weights $w^{\text{spatial}}_j, w^{\text{temporal}}_j \in [0,1]$ satisfying $w^{\text{spatial}}_j + w^{\text{temporal}}_j = 1$. The design incorporates two strategies:

\paragraph{Fixed Heuristic Gating (Baseline)}
Define weights based on branch maturity:
\begin{equation}
w^{\text{spatial}}_j =
\begin{cases}
\alpha_{\text{new}} = 0.7 & \text{if } b_j \text{ is new (no prior buds)} \\
\alpha_{\text{exist}} = 0.35 & \text{if } b_j \text{ appeared in } t-1
\end{cases}
\label{eq:fixed_gate}
\end{equation}
This prior reflects domain knowledge: new branches rely on spatial proximity (lacking temporal evidence), while mature branches leverage motion continuity (more reliable than ambiguous spatial cues in dense canopies).

\paragraph{Learnable Dynamic Gating (Advanced)}
To enable data-driven adaptation, we introduce a per-branch MLP that predicts weights from evidence statistics:
\begin{align}
\mathbf{h}_j &= [\mu^{\text{spatial}}_j, \mu^{\text{temporal}}_j, \sigma^{\text{vert}}_j, \mathbb{1}_{\text{history}}(b_j)] \in \mathbb{R}^4 \\
w^{\text{spatial}}_j &= \sigma(\text{MLP}_{\text{gate}}(\mathbf{h}_j))
\end{align}
where:
\begin{itemize}
    \item $\mu^{\text{spatial}}_j = \text{mean}_i M^{\text{spatial}}_{ij}$: Average spatial affinity across buds.
    \item $\mu^{\text{temporal}}_j = \text{mean}_i M^{\text{temporal} \to \text{branch}}_{ij}$: Average temporal affinity.
    \item $\sigma^{\text{vert}}_j$: Normalized vertical deviation (capturing motion irregularity).
    \item $\mathbb{1}_{\text{history}}(b_j) \in \{0,1\}$: Binary indicator of historical bud existence.
\end{itemize}
The MLP is initialized to replicate fixed gating behavior (bias set to logit of $\alpha_{\text{exist}}$, weights near zero), ensuring stable training onset while allowing gradient-based refinement.

\subsubsection{Score Fusion and Matching}

Fused logits are computed as a weighted combination scaled by temperature coefficients:
\begin{equation}
M^{\text{fusion}}_{ij} = w^{\text{spatial}}_j \cdot \frac{M^{\text{spatial}}_{ij}}{\tau_{\text{spatial}}} + w^{\text{temporal}}_j \cdot \frac{M^{\text{temporal} \to \text{branch}}_{ij}}{\tau_{\text{temporal}}}
\end{equation}
At inference, the Hungarian algorithm~\cite{kuhn1955hungarian} solves the optimal assignment problem on $M^{\text{fusion}}$ to enforce one-to-one bud-branch correspondence, preventing duplicates (one bud matched to multiple branches) or omissions.

\subsubsection{Loss Function}

During training, we minimize a cross-entropy loss over fused logits:
\begin{equation}
\mathcal{L} = \mathcal{L}_{\text{CE}}(M^{\text{fusion}}, y^{\text{GT}}),
\end{equation}
where $\mathcal{L}_{\text{CE}}$ is computed over $\text{softmax}(M^{\text{fusion}})$ by comparing predicted assignments to ground truth $y^{\text{GT}}$ and includes an explicit ``unmatched'' column that allows buds without valid branch assignments to be modeled. In practice, we also clamp per-branch spatial weights $w^{\text{spatial}}_j$ to lie within $[\alpha_{\min}, \alpha_{\max}]$ (e.g., $\alpha_{\min} = 0.05, \alpha_{\max} = 0.95$) to prevent degenerate collapse to a single evidence source without introducing a separate explicit regularization term. Although our implementation additionally supports an optional Hungarian matching loss that supervises the one-to-one assignment produced at inference time, this component is disabled (weight set to zero) in all reported experiments.

\subsubsection{Design Rationale}

The adaptive fusion design addresses stage-dependent failure modes:
\begin{itemize}
    \item \textbf{Early stages}: Sparse, well-separated branches $\Rightarrow$ spatial evidence dominates ($w^{\text{spatial}} \approx 0.7$).
    \item \textbf{Late stages}: Dense, overlapping branches $\Rightarrow$ temporal evidence dominates ($w^{\text{temporal}} \approx 0.7$).
    \item \textbf{New branch emergence}: Lacking historical trajectories $\Rightarrow$ fallback to spatial priors.
    \item \textbf{Occlusion events}: Bud temporarily disappears then reappears $\Rightarrow$ temporal continuity recovers identity despite spatial gap.
\end{itemize}
By learning gate weights from evidence statistics, the model automatically adjusts to unseen scenarios without manual hyperparameter tuning.

\section{Experiments}
\label{sec:experiments}

\subsection{Dataset and Preprocessing}
The multi-view temporal imagery of \textit{Brassica napus} used in this study originates from an automated greenhouse phenotyping platform~\cite{williams2023brassica} that records the full branch development trajectory from first flowering to pod fill. The dataset comprises 360 individual plants observed from three side views (0$^{\circ}$/45$^{\circ}$/90$^{\circ}$), resulting in 4,877 frames with human-verified annotations for 22,911 branch points (4.70 per frame) and 38,187 buds (7.83 per frame). This detailed labeling enables direct evaluation of branch-point-to-bud association.

Image preprocessing begins with standardizing the raw input resolutions (1100$\times$1840 or 1200$\times$1850 pixels). We first apply a static crop to remove background regions unrelated to the plant, then resize the cropped images to a uniform 640$\times$640 pixels via bilinear interpolation. For compatibility with our network architecture, these images are further resized to 224$\times$224. A binary foreground mask is generated using Li's minimum cross-entropy thresholding~\cite{li1993minimum} and concatenated with the RGB image. For temporal models, we construct an 8-channel input by stacking the 4-channel (RGB+mask) tensors from the current and previous frames, providing essential motion context.

From the annotations, we extract geometric and motion features for spatiotemporal matching. Branch points are encoded as 6-dimensional vectors $[\text{order},\allowbreak x,\allowbreak y,\allowbreak \sin\theta,\allowbreak \cos\theta,\allowbreak \text{reserved}]$, where $\theta$ denotes the initial branch orientation, while buds are represented as 5-dimensional vectors $[\text{order},\allowbreak c_x,\allowbreak c_y,\allowbreak w,\allowbreak h]$ capturing their center position and bounding box dimensions. Motion features are computed from tracked bud displacements using actual date intervals parsed from frame metadata, yielding an 11-dimensional representation: $[\text{order},\allowbreak x,\allowbreak y,\allowbreak w,\allowbreak h,\allowbreak v_x,\allowbreak v_y,\allowbreak a_x,\allowbreak a_y,\allowbreak \cos\theta,\allowbreak \sin\theta]$, where velocities and accelerations are derived via finite differences over consecutive frames. All spatial coordinates are pre-normalized to the range $[0,1]$ relative to image dimensions, providing stable anchors for matching without requiring runtime feature extraction. Crucially, although the explicit ``order'' index is retained in the raw feature vectors, it is excluded from the neural network input during forward propagation to prevent information leakage.

To prevent data leakage across developmental stages, we partition each plant's temporal sequence chronologically: the first 70\% of frames are used for training, the next 15\% for validation, and the final 15\% for testing. During training, on-the-fly augmentations are applied, including horizontal flips (50\% probability), random rotations ($\pm$10$^\circ$), and affine transformations (scale 0.95--1.05, translation $\pm$5\% of image width).

\subsection{Real Dataset Construction and Annotation System}
\label{sec:annotation_system}

While synthetic data generation enables systematic ablation studies and initial model training, robust validation and real-world deployment necessitate high-quality annotations of genuine rapeseed images captured under field conditions.
Manual annotation of slender, overlapping branch structures is notoriously labor-intensive, error-prone, and suffers from inter-annotator inconsistency, particularly in time-series datasets where the same plant individual must be tracked across developmental stages~\cite{pound2017deep}.
To address these challenges, we developed a specialized annotation system incorporating computer-vision-assisted algorithms and intelligent workflow design, significantly enhancing both annotation efficiency and structural consistency.
This section provides a concise overview of the system's core innovations, emphasizing features critical to facilitating the creation of training data for PBG-Net.

Manual annotation of time-series plant images presents unique challenges in maintaining cross-frame consistency.
Traditional annotation tools (e.g., LabelImg, CVAT) assign frame-independent IDs, which fail to preserve semantic identity of individual branches as morphology evolves over 100+ temporal frames.
When branches shift positions or undergo structural changes, conventional ID-based tracking suffers from identity drift, requiring extensive manual re-alignment.
These limitations motivated the development of a specialized annotation methodology incorporating semantic matching and computer-vision-assisted calibration.

Our annotation system introduces three technical innovations to address these challenges.
First, order-based matching replaces frame-specific IDs with semantic order indices (e.g., ``3rd lateral branch from bottom''), encoded as triplets $(\text{order}, \text{type}, \text{custom\_type\_id})$ where order represents botanical rank within an annotation category.
This semantic anchoring enables real-time synchronization: annotations added or modified in frame $t$ automatically propagate to frames $t+1$ through $T$ via order matching rather than pixel coordinates, achieving $>90\%$ reduction in redundant adjustments across time-series datasets.
Second, SIFT-based adaptive calibration automates position refinement under plant motion or perspective changes~\cite{lowe2004sift}.
For each annotated junction in frame $t$, the system extracts a local template and performs multi-scale template matching in frame $t+1$ across four graduated search radii, accepting matches with normalized cross-correlation confidence $>0.6$.
This graduated strategy balances precision with robustness to significant motion.
Third, a custom annotation type system supports flexible definition of point-based and region-based annotation categories (branch junctions, flower buds, leaf attachments), each with independent order indexing to eliminate global index conflicts when new types are added mid-project.

Validation on 50 time-series images (10 plants $\times$ 5 time points) demonstrated large improvements: annotation time reduced from 6.8 to 2.3 hours per plant (66\% reduction), inter-frame identity consistency increased from 73.5\% to 98.2\%, and SIFT calibration maintained sub-2-pixel mean absolute error versus 4.8-pixel drift in manual annotations.
The system exports annotations in two formats: Complete (full metadata) and Pure (minimal JSON).

\begin{figure*}[!t]
\centering
\includegraphics[width=\textwidth]{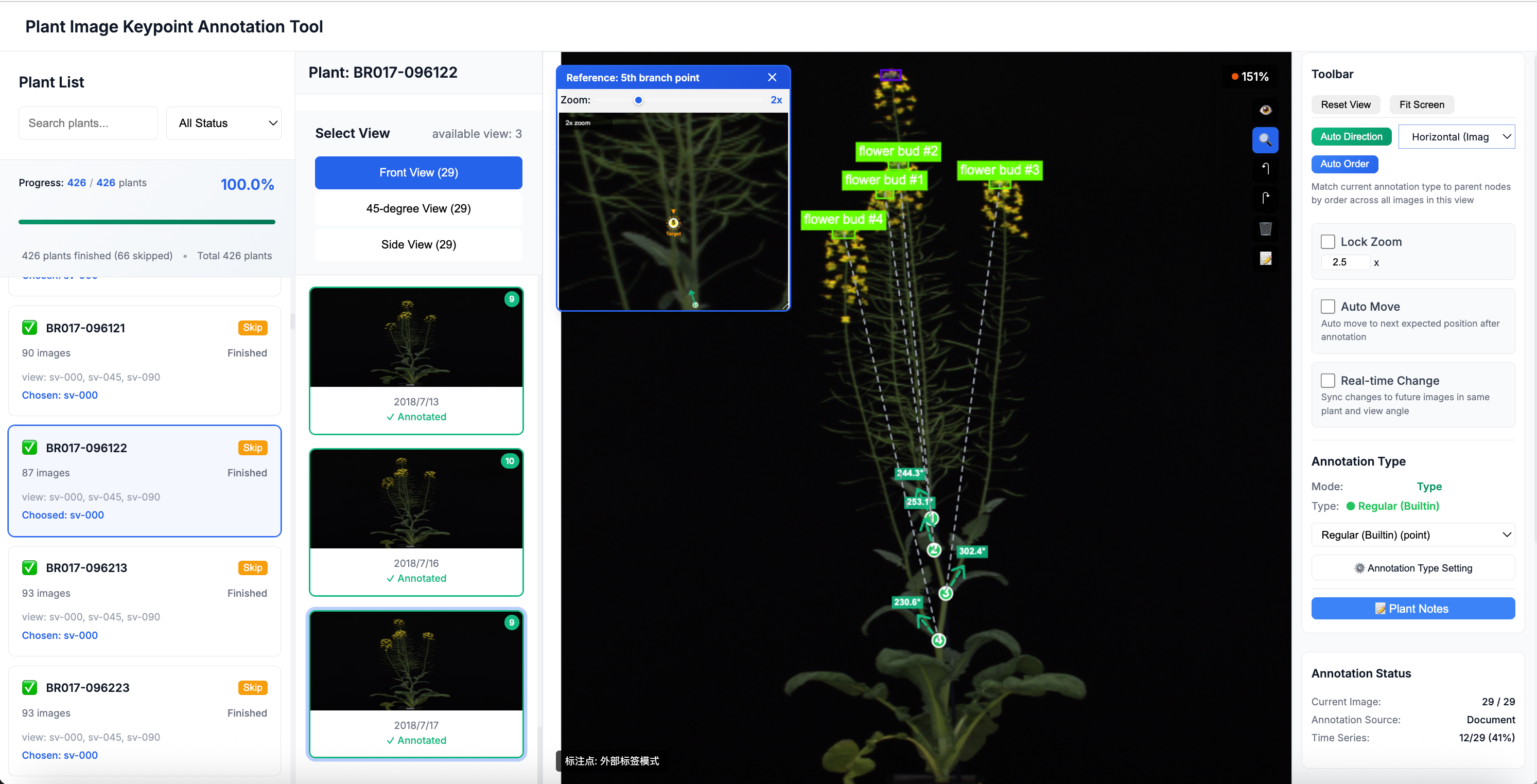}
\caption{The annotation system interface demonstrating order-based semantic matching and SIFT-based adaptive calibration. The system enables efficient tracking of branch junctions across time-series frames through intelligent workflow design and computer-vision-assisted algorithms.}
\label{fig:annotation_tool}
\end{figure*}

\subsection{Implementation Details}
All experiments instantiate the three models described in Section~\ref{sec:overall_framework} on top of a ViT-B/16 encoder: the single-frame spatial baseline, the pairwise temporal baseline with motion priors, and the dual-decoder fusion model with adaptive gating. The spatial baseline retains 4-channel patch embeddings and drives the DirectMatching decoder with 256 hidden dimensions. The temporal baseline and fusion model share 8-channel patch embeddings and expand the hidden size to 512 so that the temporal decoder can capture displacement, velocity, and acceleration cues.

We optimize every model with AdamW ($\beta_1=0.9,\beta_2=0.999$), a base learning rate of $1\times10^{-4}$, batch size 2, and gradient clipping at $\|g\|_2 \leq 1.0$. The gating and temperature parameters in the fusion model receive a dedicated $1\times10^{-3}$ learning rate scheduled by CosineAnnealingLR over 60 cycles. The spatial and temporal baselines are trained for 60 epochs under a constant learning rate to reflect their smaller parameter budgets.

To ensure reproducibility, we publish all random seeds, weight initializations, data-loader orders, and mixed-precision loss-scaling logs, which simplifies diagnosing NaNs. Inference strictly follows the procedure in Section~\ref{sec:problem_formulation}: each frame is decoded by both branches, and the fused score matrix is resolved via Hungarian matching~\cite{kuhn1955hungarian} to enforce one-to-one assignments.

\subsection{Evaluation Metrics}
We measure performance through complementary metric groups that isolate segmentation fidelity, temporal tracking stability, and branch-level structural correctness, preventing any single scalar from hiding systematic failure patterns.

\paragraph{COCO-style Instance Segmentation Metrics.} Following the COCO protocol~\cite{lin2014microsoft}, we report mean Average Precision (mAP) computed as the mean of the precision-recall integrals over IoU thresholds $0.5{:}0.05{:}0.95$, and AP50, the Average Precision evaluated at IoU $=0.5$. These metrics summarize region-level localization and classification quality for bud instances under progressively stricter spatial agreements.

\paragraph{MOT-style Tracking Metrics.} For multi-object tracking performance we adopt the MOTChallenge suite~\cite{milan2016mot16}. Multi-Object Tracking Accuracy (MOTA) jointly penalizes missed targets, false alarms, and identity changes:
\begin{equation}
\text{MOTA} = 1 - \frac{\text{FN} + \text{FP} + \text{IDSW}}{\text{GT}},
\end{equation}
where GT is the total number of annotated buds, FN counts missed detections, FP counts false positives, and IDSW (reported as ID in the tables) counts identity switches. Identity F1-score (IDF1) measures agreement between predicted and ground-truth identity trajectories,
\begin{equation}
\text{IDF1} = \frac{2 \cdot \text{IDTP}}{2 \cdot \text{IDTP} + \text{IDFP} + \text{IDFN}},
\end{equation}
with IDTP/IDFP/IDFN denoting true/false positives/negatives under identity-aware matching. We additionally report Mostly Tracked (MT, trajectories recovered for $\geq80\%$ of their lifespan) and Mostly Lost (ML, $\leq20\%$ covered) ratios, as well as the raw FP, FN, and ID counts to expose the individual error sources.

\paragraph{Branch-Specific Metrics.} To capture branch topology fidelity we retain the task-tailored metrics. \textbf{Branch Matching Accuracy (BMA)} measures the proportion of buds whose Hungarian-matched assignment aligns with the annotated branch index $y_i$:
\begin{equation}
\text{BMA} = \frac{1}{\sum_t |\mathcal{F}_t|} \sum_t \sum_{i} \mathbb{1}\big[\pi_t(f_i) = b_{y_i}\big].
\end{equation}
\textbf{Bud Localization Error (BLE)} quantifies spatial drift by averaging Euclidean distances between matched bud coordinates:
\begin{equation}
\text{BLE} = \frac{1}{\sum_t |\mathcal{F}_t|} \sum_t \sum_{i} \big\| \mathbf{p}(f_i) - \mathbf{p}(\hat{f}_i) \big\|_2.
\end{equation}
\textbf{Longitudinal Intersection over Union (LIoU)} evaluates branch skeleton length fidelity by computing the ratio between the minimum and maximum skeleton lengths extracted from ground-truth and predicted masks:
\begin{equation}
\text{LIoU} = \frac{\min(L_{\text{gt}}, L_{\text{pred}})}{\max(L_{\text{gt}}, L_{\text{pred}})},
\end{equation}
where $L$ denotes the skeleton length obtained by morphological skeletonization followed by graph-traversal length computation. This metric isolates structural elongation accuracy independent of spatial alignment.
\textbf{Branch Temporal Consistency (BTC)} measures endpoint localization precision by computing the bidirectional average nearest-neighbor distance (in pixels) between predicted and ground-truth skeleton endpoints:
\begin{equation}
\begin{split}
\text{BTC} = \frac{1}{2}\bigg( &\frac{1}{|E_{\text{pred}}|}\sum_{e \in E_{\text{pred}}} \min_{e' \in E_{\text{gt}}} \|e - e'\|_2 \\
&+ \frac{1}{|E_{\text{gt}}|}\sum_{e' \in E_{\text{gt}}} \min_{e \in E_{\text{pred}}} \|e' - e\|_2 \bigg),
\end{split}
\end{equation}
where $E_{\text{pred}}$ and $E_{\text{gt}}$ denote the sets of skeleton endpoints (pixels with $\leq1$ neighbor in the 8-connected skeleton). Lower BTC indicates more accurate branch tip positioning.
All metrics are computed independently for each camera view and growth phase (early, mid, late) and then aggregated via sample-count weighting to avoid dominance by any single regime.

\subsection{Quantitative Results}
Table~\ref{tab:main_results} presents a comprehensive quantitative evaluation on our test set, positioning ST-DETrack against state-of-the-art instance segmentation models and tracking baselines. Our proposed method establishes a new performance benchmark, achieving a Branch Matching Accuracy (BMA) of 93.6\% and a Multiple Object Tracking Accuracy (MOTA) of 0.980. Notably, ST-DETrack outperforms the spatial baseline (64.7\% BMA) by a substantial margin of 28.9 percentage points. This disparity underscores the inadequacy of relying solely on geometric priors in entangled environments, where phenotypic similarities between adjacent branches create severe spatial ambiguity. Furthermore, the method surpasses the temporal baseline (90.3\% BMA) by 3.3 percentage points. While the temporal baseline is robust in dense stages, it falters with newly emerged branches that lack motion history; our dual-decoder fusion effectively resolves this by leveraging spatial cues during these initialization phases.

\begin{figure*}[!t]
  \centering
  \includegraphics[width=0.95\textwidth,height=0.85\textheight,keepaspectratio]{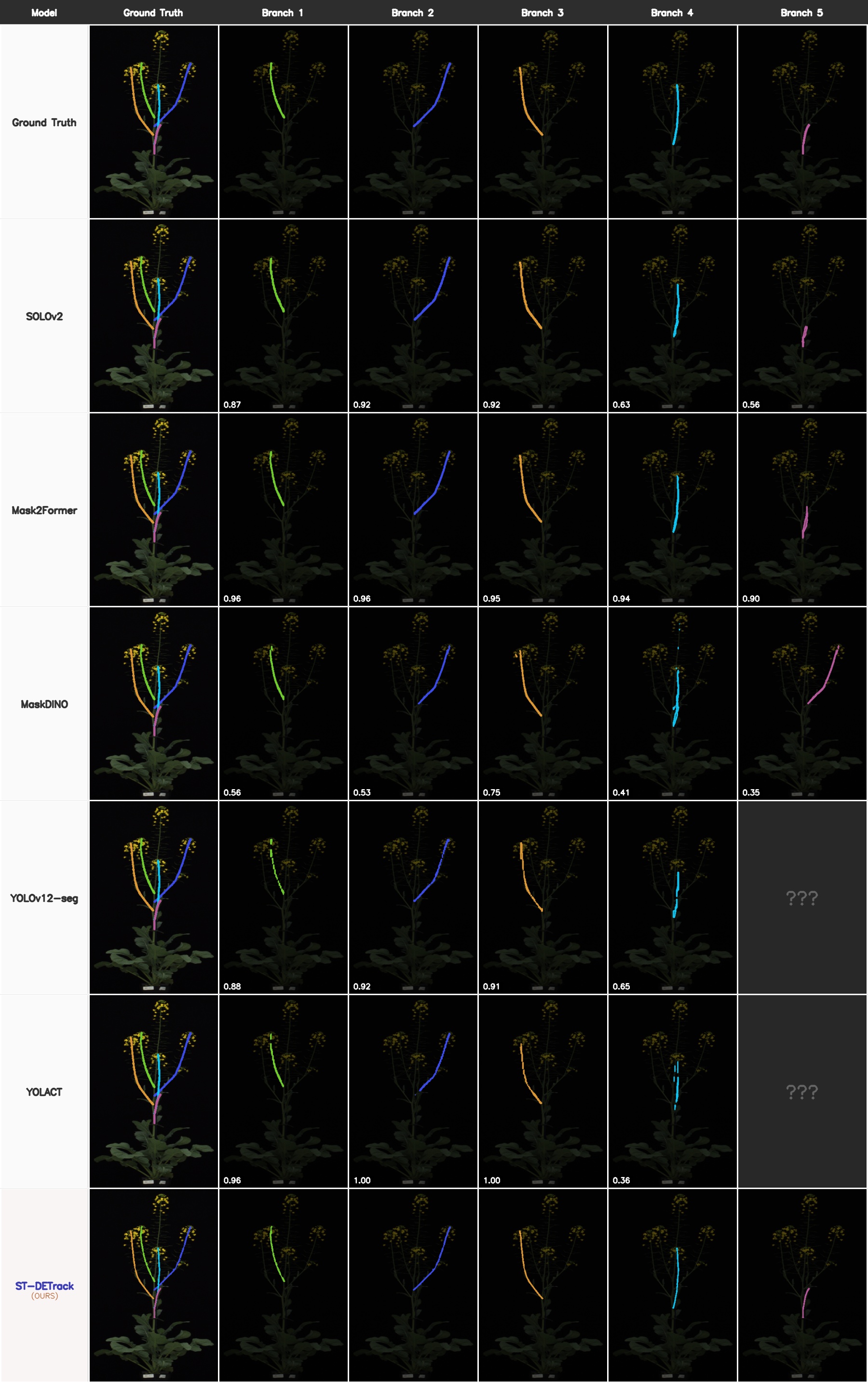}
  \caption{Qualitative comparison of instance segmentation models on branch-level predictions. Each row shows predictions from a different model (Ground Truth, SOLOv2, Mask2Former, MaskDINO, YOLOv12-seg, YOLACT, and ST-DETrack) for the same plant sample. Columns represent individual branches (Branch 1--5) with their corresponding IoU scores displayed below each prediction. Color-coded masks indicate different branch instances. Ground truth annotations (top row) serve as the reference for evaluation. Note that while SOLOv2, Mask2Former, and YOLOv12-seg achieve relatively high IoU scores on well-separated branches, they struggle with fine-grained boundary delineation. MaskDINO and YOLACT show inconsistent performance with lower IoU values, particularly on occluded or thin branches. ST-DETrack (bottom row) demonstrates superior segmentation quality across all branches by leveraging spatiotemporal evidence, achieving more accurate boundary localization and better handling of challenging cases such as inter-branch occlusions and morphological variations. The "???" markers indicate failed detections where the model completely missed the branch instance.}
  \label{fig:model_comparison}
\end{figure*}

Comparisons with static instance segmentation models (e.g., Mask R-CNN, Mask2Former, YOLOv12-seg) highlight the critical importance of temporal continuity in phenotyping. While advanced transformers like Mask2Former achieve a competitive LIoU of 0.8233, they suffer from significant temporal inconsistency, evidenced by a Branch Temporal Consistency (BTC) of 20.75 pixels. In sharp contrast, ST-DETrack achieves a perfect BTC of 0, indicating absolute endpoint consistency across all test sequences. The catastrophic failure of generic MOT trackers (TrackFormer: MOTA $-6.753$; TransTrack: MOTA $-0.216$) reveals a fundamental domain mismatch. These trackers rely on constant-velocity motion priors designed for rigid objects (e.g., pedestrians), which are incompatible with the non-rigid, sessile growth patterns of plants. Our results validate that the proposed negative gravitropism prior and spatiotemporal fusion are essential for handling the complex morphological metamorphosis of plant growth. For visual comparison of segmentation quality, Figure~\ref{fig:model_comparison} illustrates representative branch-level predictions from different models.

\begingroup
\renewcommand{\arraystretch}{2.0}  
\begin{table*}[!t]
    \centering
    \caption{Test set results for instance segmentation and tracking evaluation. We report COCO-style mAP and AP50 for segmentation models, MOT-style metrics (MOTA, IDF1, FP, FN) for tracking models, and branch-specific metrics (BMA, LIoU, BTC) tailored for botanical structure analysis. LIoU: Longitudinal Intersection over Union; BTC: Branch Temporal Consistency (pixel error). Higher is better for mAP, AP50, BMA, LIoU, MOTA, and IDF1. Lower is better for FP, FN, and BTC.}
    \label{tab:main_results}
    \small
    \begin{tabular}{>{\centering\arraybackslash}p{3.5cm} >{\centering\arraybackslash}p{1.35cm} >{\centering\arraybackslash}p{1.50cm} >{\centering\arraybackslash}p{1.30cm} >{\centering\arraybackslash}p{1.0cm} >{\centering\arraybackslash}p{0.85cm} >{\centering\arraybackslash}p{0.85cm} >{\centering\arraybackslash}p{1.0cm} >{\centering\arraybackslash}p{1.20cm} >{\centering\arraybackslash}p{1.0cm}}
        \hline
        & \multicolumn{2}{c}{\textbf{Instance Seg}} & \multicolumn{4}{c}{\textbf{Multi-Object Tracking (MOT)}} & \multicolumn{3}{c}{\textbf{Branch-Specific}} \\
        \textbf{Model} & \textbf{F1@0.5 $\uparrow$} & \textbf{mAP50 $\uparrow$} & \textbf{MOTA $\uparrow$} & \textbf{IDF1 $\uparrow$} & \textbf{FP $\downarrow$} & \textbf{FN $\downarrow$} & \textbf{BMA $\uparrow$} & \textbf{LIoU $\uparrow$} & \textbf{BTC $\downarrow$} \\
        \hline
        Spatial Baseline & -- & -- & -- & -- & -- & -- & 0.647 & -- & -- \\
        Temporal Baseline & -- & -- & 0.957 & 0.978 & \textbf{0} & 23 & 0.903 & -- & -- \\
        \textbf{ST-DETrack (Ours)} & -- & -- & \textbf{0.980} & \textbf{0.993} & \textbf{0} & \textbf{5} & \textbf{0.936} & \textbf{0.9617} & \textbf{0} \\
        \hline
        Mask R-CNN~\cite{he2017mask} & 0.420 & 0.384 & -- & -- & -- & -- & -- & 0.5328 & 132.20 \\
        Mask2Former~\cite{cheng2022mask2former} & 0.778 & 0.740 & -- & -- & -- & -- & -- & 0.8233 & 20.75 \\
        MaskDINO~\cite{li2022maskdino} & 0.522 & 0.632 & -- & -- & -- & -- & -- & 0.7293 & 40.23 \\
        YOLOv12-seg~\cite{tian2025yolov12} & 0.683 & 0.645 & -- & -- & -- & -- & -- & 0.6443 & 122.69 \\
        SOLOv2~\cite{solov2020} & 0.458 & 0.710 & -- & -- & -- & -- & -- & 0.8380 & 23.65 \\
        YOLACT~\cite{yolact2019} & 0.374 & 0.316 & -- & -- & -- & -- & -- & 0.6560 & 122.90 \\
        PBG-Net & 0.565 & 0.406 & -- & -- & -- & -- & -- & 0.6553 & 2.30 \\
        \hline
        TrackFormer~\cite{meinhardt2022trackformer} & -- & -- & -6.753 & 0.002 & 22888 & 3360 & -- & -- & -- \\
        TransTrack~\cite{sun2020transtrack} & -- & -- & -0.216 & 0.002 & 733 & 3385 & -- & -- & -- \\
        \hline
    \end{tabular}
\end{table*}
\endgroup

\subsection{Qualitative Analysis}
\begin{figure*}[t]
  \centering
  \includegraphics[width=0.95\textwidth]{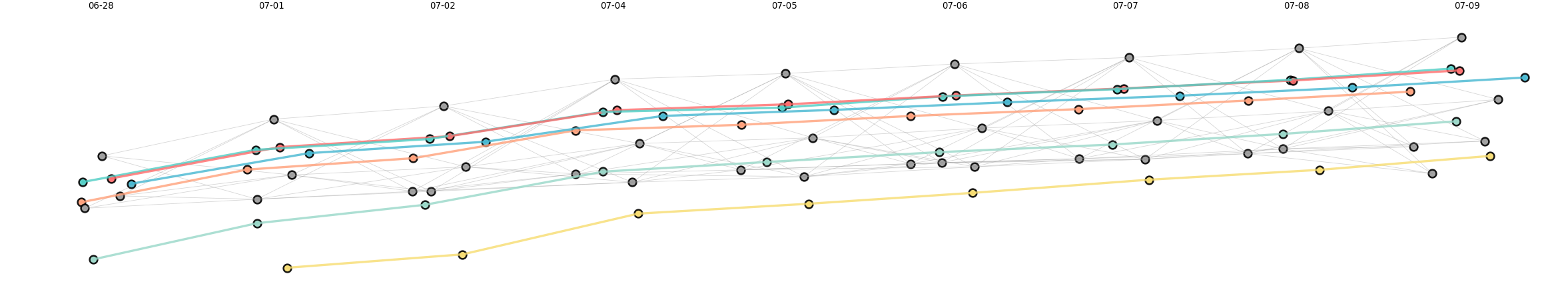}\\[0.3em]
  \includegraphics[width=0.95\textwidth]{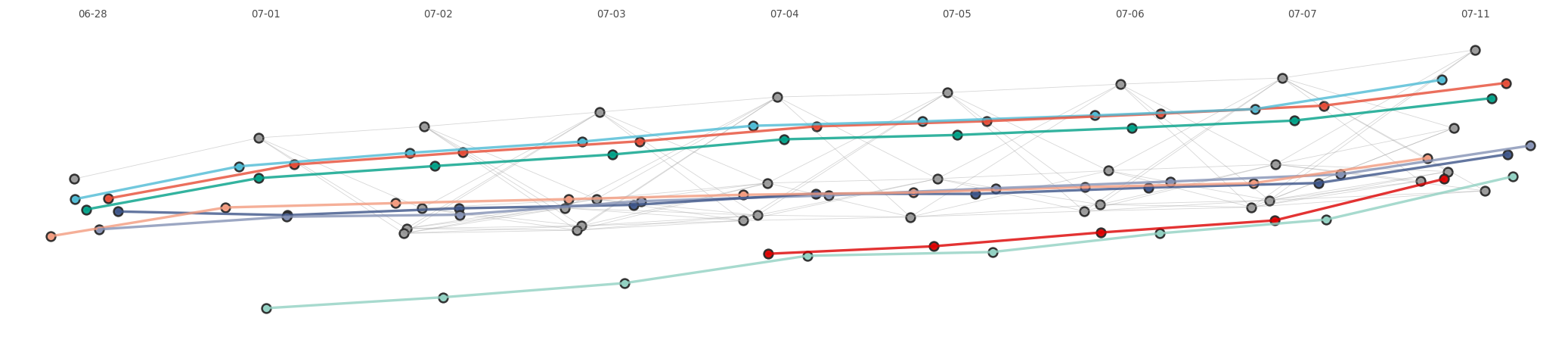}\\[0.3em]
  \includegraphics[width=0.95\textwidth]{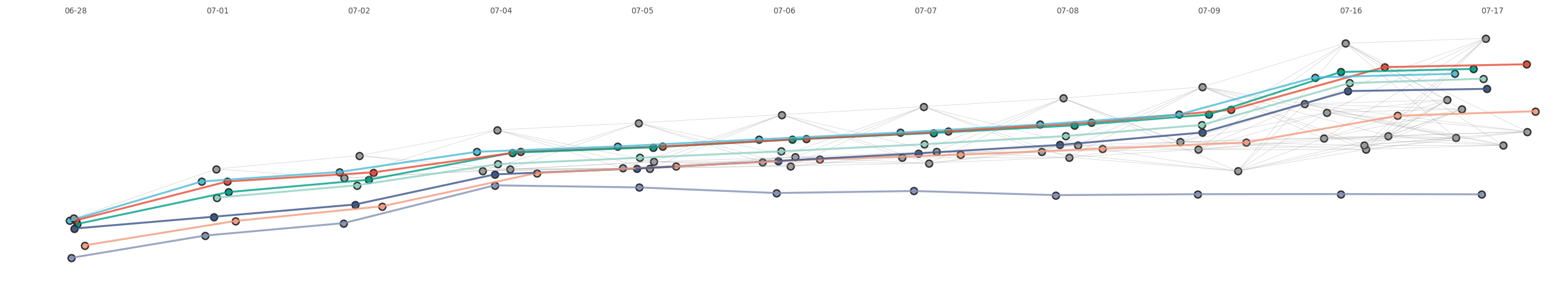}\\[0.3em]
  \includegraphics[width=0.95\textwidth]{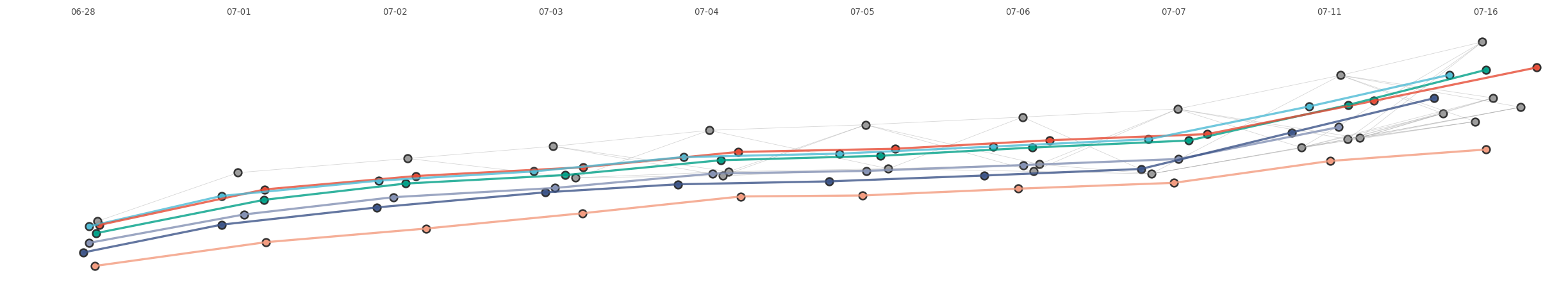}\\[0.3em]
  \includegraphics[width=0.95\textwidth]{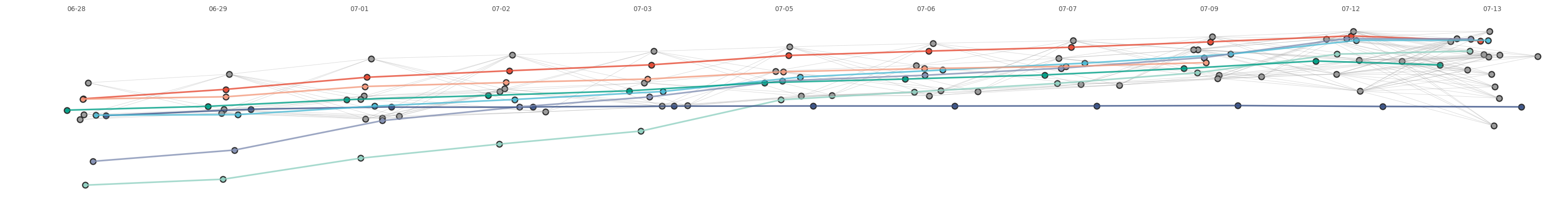}
  \caption{Longitudinal branch identity tracking visualization across five representative test plants. Each row shows a temporal sequence (left to right: progressive time steps) for a single plant with color-coded branch IDs maintained consistently across frames. From top to bottom: (a) BR017-028111 and (b) BR017-028112 demonstrate early-stage sparse configurations with well-separated branches; (c) BR017-028121 and (d) BR017-028122 show mid-stage transitions with increasing spatial overlap; (e) BR017-028213 exhibits late-stage dense canopy with severe inter-branch occlusions. Individual primary branches retain unique colors throughout their development, validating the spatiotemporal fusion mechanism's ability to handle diverse structural complexity and growth stages. The horizontal arrangement within each row demonstrates temporal consistency, where identical colors across time steps confirm successful identity preservation.}
  \label{fig:longitudinal_tracking}
\end{figure*}

Figure~\ref{fig:longitudinal_tracking} visualizes the longitudinal tracking performance of ST-DETrack across complete growth cycles for five representative test plants (BR017-028111 through BR017-028213). The visualization spans from early vegetative stages to late flowering, with individual primary branches color-coded to denote identity. In the early stages (left columns), where branches are sparse but spatially erratic, the model correctly distinguishes between distinct growth points, overcoming the spatial ambiguity that typically confuses proximity-based methods. As the sequences progress to late-stage dense canopies (right columns), the branches intertwine extensively. Despite severe occlusions and morphological overlap, the consistent color mapping confirms that the model preserves identity without switching, corroborating the low False Negative rate (FN=5) and high IDF1 (0.993) reported quantitatively.

The qualitative results also illustrate the model's dynamic adaptability to stage-dependent challenges. For newly emerged branches, which present a ``cold start'' problem for temporal trackers, the visualization shows successful initialization and stable tracking after a brief warm-up period ($\sim$2--3 frames), attributed to the structural guidance of the spatial decoder. Conversely, in mature stages where spatial boundaries blur, the temporal decoder's motion coherence ensures trajectory smoothness. This visual evidence aligns with the quantitative metrics, particularly the zero BTC score, demonstrating that ST-DETrack effectively bridges the gap between spatial segmentation and temporal association to deliver robust, coherent, and identity-preserving tracking in complex plant phenotyping scenarios.

\section{Discussion}
\label{sec:discussion}

\subsection{Effectiveness of Spatiotemporal Fusion}
The quantitative results in Table~\ref{tab:main_results} strongly validate our hypothesis that fusing spatial and temporal evidence is essential for robust branch tracking. The proposed Fusion model achieves a Branch Matching Accuracy (BMA) of \textbf{93.6\%}, significantly outperforming the Spatial Baseline (64.7\%) by \textbf{28.9 percentage points} and the Temporal Baseline (90.3\%) by \textbf{3.3 percentage points}.

This performance differential highlights the stage-dependent nature of the problem. The Spatial Baseline relies solely on geometric proximity. While effective in early growth stages where branches are sparse, it collapses in late-stage dense canopies where average inter-bud distances shrink below the ambiguity threshold. Conversely, the Temporal Baseline excels in these dense scenarios by leveraging motion continuity but struggles with branch initiation---newly emerged buds lack the historical trajectory required for temporal association. Our adaptive gating mechanism successfully bridges these regimes, dynamically shifting weight to spatial priors for new branches ($w^{\text{spatial}} \approx 0.7$) and to temporal priors for mature, entangled branches, thereby resolving the "stage-dependent ambiguity" defined in Section~\ref{sec:intro}.

\subsection{Comparison with Existing Vision Frameworks}
Standard computer vision baselines struggle to adapt to the biological constraints of plant growth.

\textbf{Static Instance Segmentation Limitations.} Methods like Mask R-CNN~\cite{he2017mask} and SOLOv2~\cite{solov2020} treat each frame independently. Although they can segment disjoint organs, they fail to maintain identity consistency over time. As shown in Table~\ref{tab:main_results}, Mask R-CNN achieves a modest mAP50 of 0.384. In a longitudinal context, this manifests as \textit{identity fragmentation}: a single growing branch is assigned dozens of different IDs over its lifespan due to minor morphological changes, rendering the output useless for trait trajectory analysis.

\textbf{Generic Multi-Object Tracking Limitations.} Transformer-based trackers such as TrackFormer~\cite{meinhardt2022trackformer} and TransTrack~\cite{sun2020transtrack} perform poorly on this task, yielding negative MOTA scores (-6.75 and -0.22, respectively). This catastrophic failure stems from fundamental mismatching of motion priors. These models typically assume inertial motion (constant velocity or momentum) suitable for rigid bodies like pedestrians or vehicles. Plants, however, exhibit \textit{non-rigid elongation} driven by biological growth rather than momentum. Furthermore, the "birth rate" of new objects (buds) in plants is far higher than in typical MOT datasets, confusing query-propagation mechanisms designed for stable object counts. Our approach replaces these inertial assumptions with growth-centric priors, resulting in stable tracking (MOTA 0.980).

\subsection{Impact of Biological Priors}
A key distinguishing feature of ST-DETrack is the incorporation of domain-specific biological constraints, specifically the \textbf{Negative Gravitropism Prior}. By enforcing that vertical displacement must align with the plant's global upward growth ($\Delta y_{ik} \approx \Delta y_{\text{global}}$), we effectively prune physically impossible associations. For instance, in dense clusters where a lower branch's bud is spatially close to an upper branch's bud from the previous frame, a pure Euclidean matcher might incorrectly associate them. The gravitropic constraint penalizes this "downward" match, guiding the model toward the biologically correct "upward" trajectory. This is distinct from generic motion smoothing and is critical for distinguishing interwoven branches in 2D projection.

\subsection{Implications for High-Throughput Phenotyping}
The transition from static to dynamic phenotyping is a bottleneck in modern plant science. The high Branch Matching Accuracy (93.6\%) achieved by our system enables the reliable extraction of \textit{longitudinal traits}---such as branch elongation rates, flowering duration, and angle changes over time---from large-scale populations. Unlike static traits (e.g., final yield), these dynamic traits provide deeper insights into the genetic control of developmental timing (phenology). By automating the extraction of individual branch trajectories, ST-DETrack paves the way for Genome-Wide Association Studies (GWAS) on structural plasticity, potentially identifying gene loci that optimize canopy architecture for light interception and mechanical harvesting.

\subsection{Future Directions}
Several avenues warrant further investigation to expand the method's scope and robustness. In the short term, eliminating the dependency on manual branch point annotations through end-to-end joint detection and tracking would significantly enhance practical deployability. A two-stage pipeline---first detecting junction keypoints via heatmap regression~\cite{carion2020detr}, then performing spatiotemporal matching conditioned on detected junctions---could preserve the benefits of explicit spatial anchoring while automating the annotation bottleneck. Alternatively, reformulating the problem as joint optimization ($\mathcal{L}_{\text{total}} = \mathcal{L}_{\text{detection}} + \lambda \mathcal{L}_{\text{matching}}$) might enable end-to-end gradient flow, though preliminary experiments suggest that coupling detection errors propagate into matching failures, necessitating careful regularization.

Incorporating 3D spatial information represents another promising direction. Future work could explore integrating depth measurements, such as those from LiDAR or stereo cameras, to disambiguate branches that overlap in 2D projection but are spatially separated in depth. Extending the spatiotemporal fusion framework to a spatiotemporal-3D model---where queries encode $(x, y, z, t)$ rather than $(x, y, t)$---may resolve ambiguities that neither spatial nor temporal evidence alone can address. This extension is particularly relevant for multi-view configurations: if future hardware supports synchronized multi-camera imaging, 3D reconstruction could unify cross-view correspondences into a coherent 3D+time representation.

Longer-term research directions include evaluating cross-species transferability and scaling to whole-season phenotyping. The current framework has been validated exclusively on \textit{Brassica napus}; whether the learned spatiotemporal priors generalize to crops with divergent branching architectures (e.g., maize with predictable phyllotaxis versus tomato with indeterminate lateral meristems) remains an open question. Transfer learning experiments, where the fusion model is pretrained on rapeseed then fine-tuned on small target-crop datasets, could assess the universality of our architectural design. Additionally, extending temporal windows from 50+ frames to 100+ days (covering entire growth cycles from germination to senescence) will require addressing long-term appearance drift and handling topological events such as branch senescence and abscission. Incorporating re-identification modules---commonly used in long-term person tracking~\cite{wojke2017simple}---may enable recovery from prolonged occlusions or identity confusion spanning weeks.

Finally, real-time deployment on edge devices for in-field phenotyping presents both engineering challenges and opportunities. Our current implementation processes batches offline; adapting the framework for online inference would enable reactive phenotyping systems that trigger interventions (e.g., targeted irrigation, pest control) based on detected developmental anomalies. Model compression techniques (quantization, pruning) and efficient Transformer variants (e.g., lightweight attention mechanisms) could reduce latency and memory footprint, enabling deployment on agricultural robots or edge servers in resource-constrained field settings.
\section{Conclusion}
\label{sec:conclusion}
In this work, we presented ST-DETrack, a novel spatiotemporal fusion framework for extracting individual plant branching architectures from time-series imagery. By addressing the critical challenge of ``stage-dependent ambiguity,'' our approach dynamically integrates spatial geometric priors and temporal motion evidence through an adaptive dual-decoder mechanism. We further introduced a biological constraint based on negative gravitropism, which proved essential for resolving complex entanglements in dense canopies where traditional computer vision assumptions typically fail.

Experimental results on a large-scale \textit{Brassica napus} dataset demonstrate that ST-DETrack achieves a Branch Matching Accuracy of 93.6\%, significantly outperforming state-of-the-art static segmentation models and generic multi-object trackers. The system effectively handles the non-rigid growth dynamics and severe occlusions characteristic of plant development, maintaining identity consistency from early budding to late flowering stages.

This advancement bridges the gap between raw image data and biological insight, enabling the high-throughput quantification of dynamic phenotypic traits such as branch elongation rates and structural plasticity. Such capabilities are pivotal for next-generation genomic studies and precision breeding programs aimed at optimizing crop architecture for yield and environmental resilience. Future work will focus on extending this framework to end-to-end joint detection and tracking, as well as exploring cross-species generalization to broader agricultural applications.

\textit{Supplementary Material}: Additional details on our preliminary work (PBG-Net baseline architecture), synthetic data generation methodology, implementation specifics, ablation studies, and qualitative results are provided in the supplementary material.

\section*{Acknowledgment}

The authors would like to thank Fiona Corke, who helped to generate the original images, Rachel Wells and Judith Irwin, who helped design the original experiment. The images were funded by the Biotechnology and Biological Sciences Research Council grant ‘Brassica Rapeseed And Vegetable Optimisation strategic Longer and Larger fund’ (BRAVO sLOLA) (BB/P003095/1).

\ifCLASSOPTIONcaptionsoff
  \newpage
\fi



%

\clearpage
%

\begin{IEEEbiography}[{\includegraphics[width=1in,height=1.25in,clip,keepaspectratio]{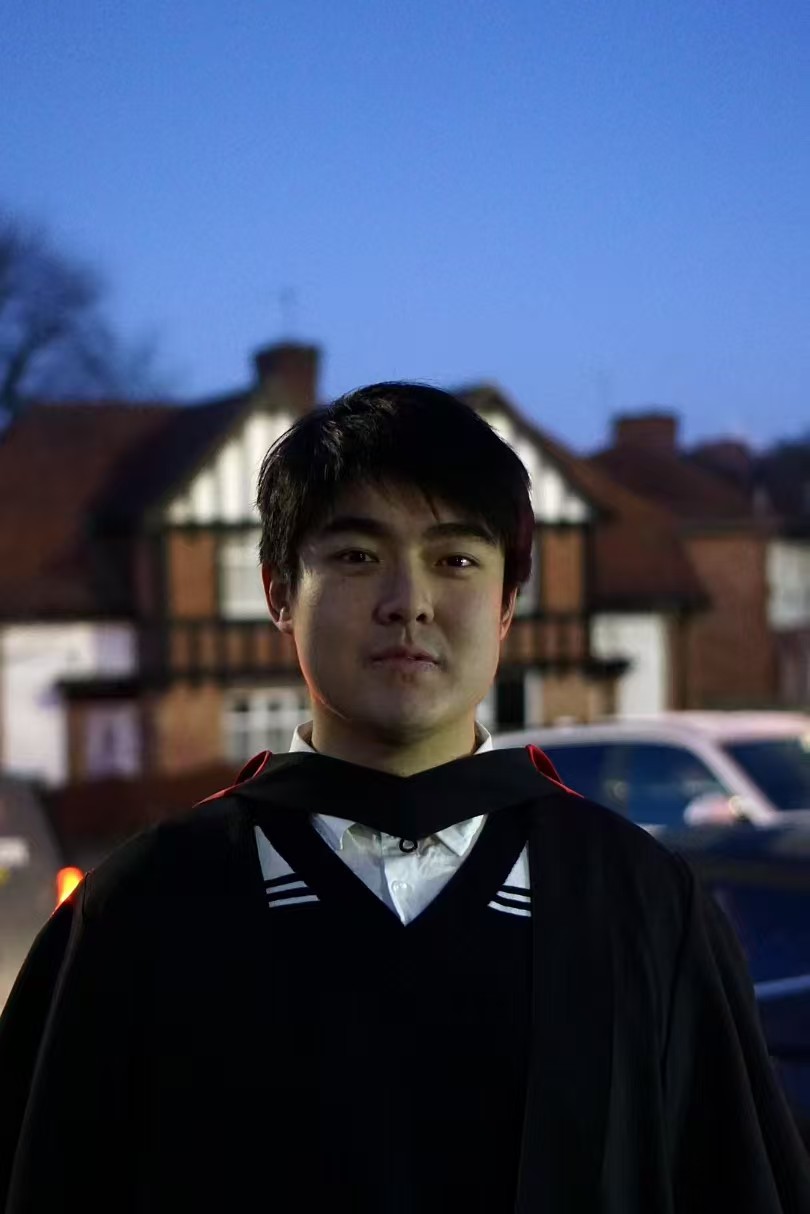}}]{Yueqianji Chen}
  received the M.S. degree in data science with Distinction from Durham University, Durham, U.K., in 2024, where he received the Outstanding Achievement Award in Data Science.
  
  He has extensive experience in entrepreneurship and software development. He served as a Technical Co-Founder of Shanxi Qingshu Network Technology Co., Ltd., participating in the research and development of multiple projects. He has also received numerous awards in various competitions. His research interests include multimodal deep learning, computer vision, and plant phenomics.
  \end{IEEEbiography}

  \begin{IEEEbiography}[{\includegraphics[width=1in,height=1.25in,clip,keepaspectratio]{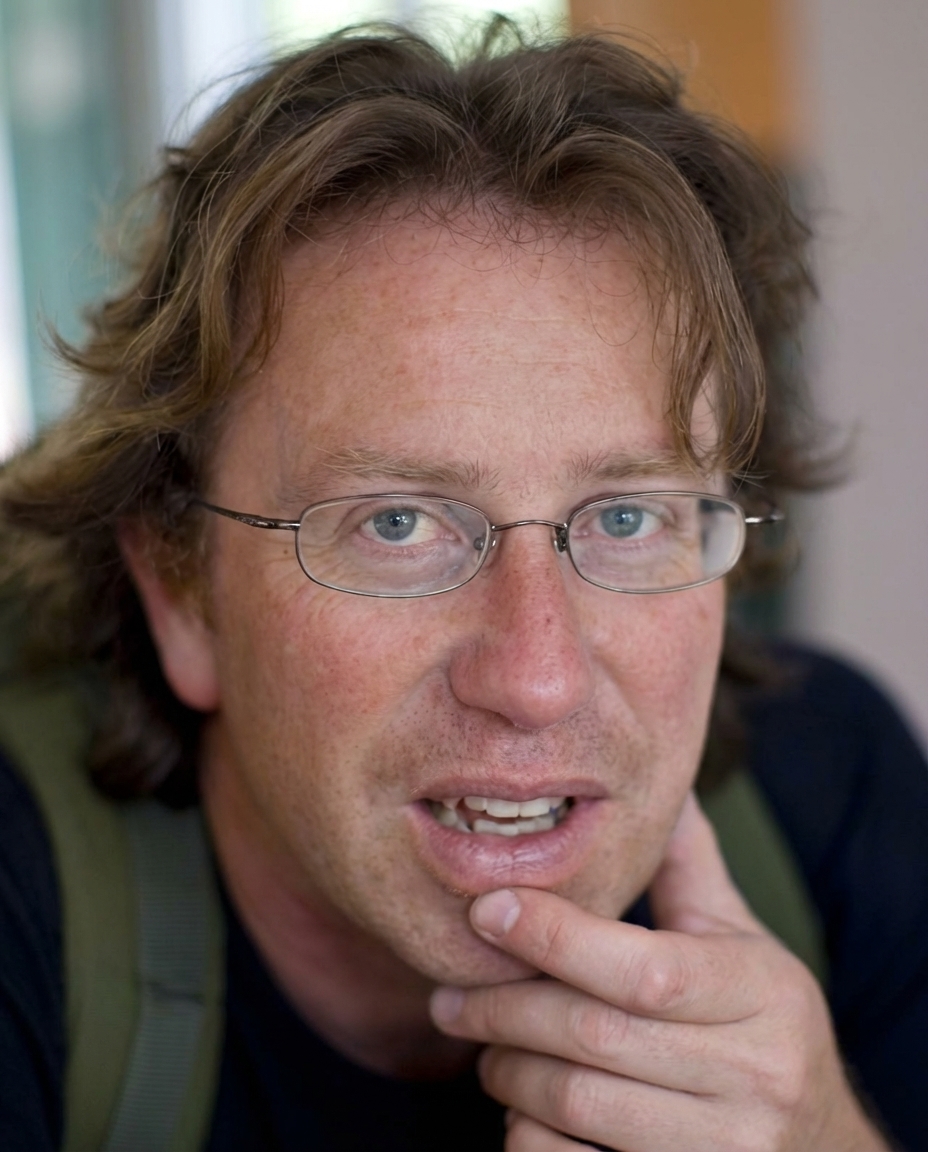}}]{Kevin Williams}
  worked for 23 years in the oil industry, picking up a couple of part-time degrees in chemistry and applied mathematics along the way, before 'retiring' to the Welsh coast in Aberystwyth. A Ph.D. (Active Learning for Drug Discovery) in a previously esoteric corner of the AI universe followed to keep the grey cells active.
  
  For the last few years, Kevin has worked as part of a very small team at the National Plant Phenomics Centre in Gogerddan, near Aberystwyth, developing methods for large plant phenotyping and data analysis thereof. Much of the work done recently at the NPPC has centred on investigating cereal and brassica genotypes, and Kevin's remit has included a broad range of projects including (but most definitely not limited to) rig-building, image analysis, crop growth models, and definition of novel digital phenotypes.
  \end{IEEEbiography}
  
  \begin{IEEEbiography}[{\includegraphics[width=1in,height=1.25in,clip,keepaspectratio]{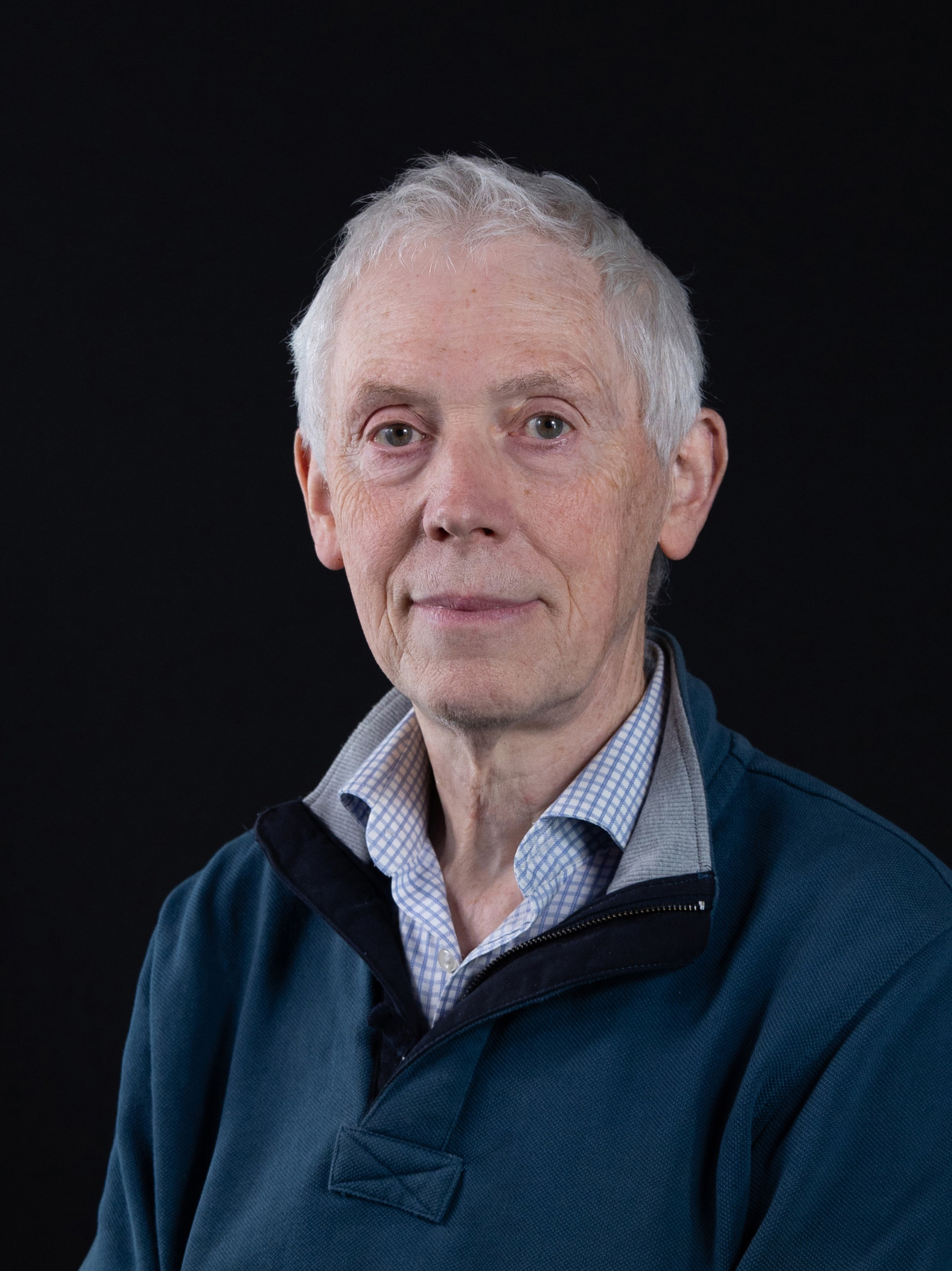}}]{John H. Doonan}
    is currently Professor of Genetics in the Institute of Biological, Environmental and Rural Sciences at Aberystwyth University and is Director of the National Plant Phenomics Centre, a BBSRC funded national facility. His research interests include the application of remote sensing, computer vision and machine learning to plant genetics and crop biology. He has published more than 100 peer-reviewed papers in leading international journals.
    \end{IEEEbiography}
    
  \begin{IEEEbiography}[{\includegraphics[width=1in,height=1.25in,clip,keepaspectratio]{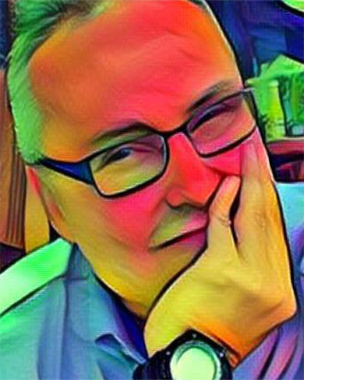}}]{Paolo Remagnino}
  (SM IEEE and AAAI) has been a full Professor in Computer Science since 2012. He joined Durham University in 2022.
  
  He is the author of over 200 articles published at international conferences and high impact scientific journals and leads research primarily on the development of innovative methods for image and video interpretation, making wide use of pattern recognition, machine and deep learning and artificial intelligence techniques. Three strands of research are currently investigated: medical image understanding, computational botany and machine learning, with particular emphasis on deep learning techniques.
  \end{IEEEbiography}
  
  \begin{IEEEbiography}[{\includegraphics[width=1in,height=1.25in,clip,keepaspectratio]{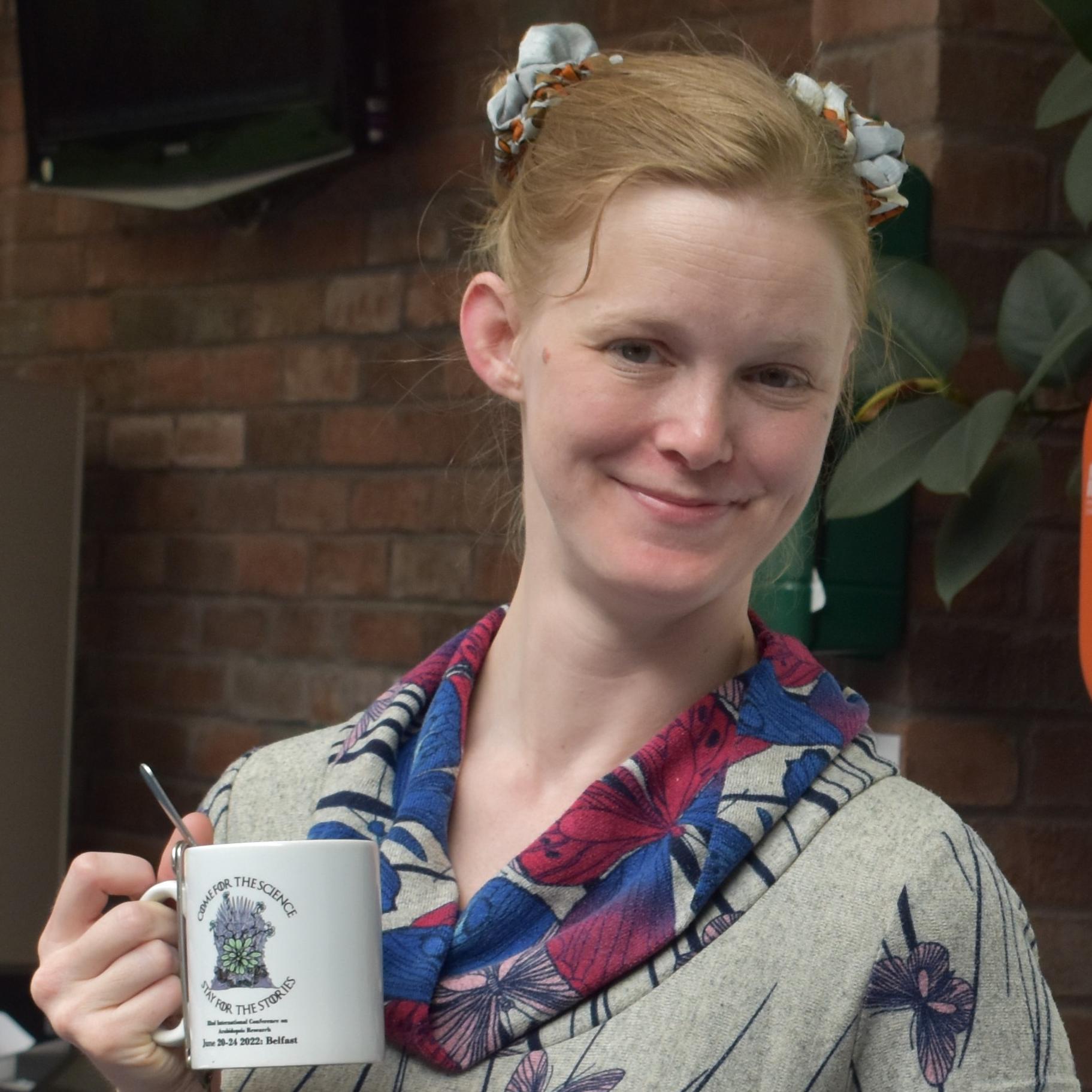}}]{Jo Hepworth}
  is a plant developmental geneticist at Durham University, U.K. Her research focuses on how plants sense and integrate environmental information, particularly temperature cues, to make developmental decisions about whole-plant morphology. She works with the model plant \textit{Arabidopsis thaliana} and \textit{Brassica} crops, employing multidisciplinary approaches including genetics, imaging, and mathematical modelling. She has published extensively in leading journals including \textit{Nature Communications}, \textit{Cell Systems}, \textit{eLife}, and \textit{New Phytologist}.
  \end{IEEEbiography}




\end{document}